\documentclass{amia}
\usepackage{graphicx}
\usepackage[labelfont=bf]{caption}
\usepackage[superscript,nomove]{cite}
\usepackage{color}
\usepackage{latexsym}
\usepackage{url}
\usepackage{graphicx,subfig}
\usepackage{booktabs}
\usepackage{tabularx}
\usepackage{changepage}
\usepackage{color,soul}
\usepackage{multirow}
\usepackage{amsmath,amssymb}
\usepackage{algorithm}
\usepackage[noend]{algpseudocode}
\usepackage{comment}
\usepackage[
  colorinlistoftodos, %enable a coloured square in the list of todos
  textwidth=\marginparwidth, %set the width of the todonotes
  textsize=scriptsize, %size of the text in the todonotes
  ]{todonotes}
  
\usepackage{xcolor,colortbl}

\begin{document}

%\title{Clinical Document Classification Using Limited Training Data: Leveraging Labeled and Unlabeled Data Across Hospitals}

\title{Clinical Document Classification Using Labeled and Unlabeled Data Across Hospitals}

% Clinical Document Classification Using Small Data: Leveraging Labeled and Unlabeled Data Across Hospitals

% Using Labeled and Unlabeled Data Across Hospitals for Clinical Document Classification

\author{Hamed Hassanzadeh, PhD$^{1}$, Mahnoosh Kholghi, PhD$^{1}$, Anthony Nguyen, PhD$^{1}$, Kevin Chu, MBBS FACEM$^{2}$}

\institutes{
    $^1$Australian e-Health Research Centre, CSIRO, Brisbane, QLD, Australia; \\ %$^2$Queensland University of Technology, Brisbane, QLD, Australia; \\ 
    $^2$Royal Brisbane and Women’s Hospital, Brisbane, QLD, Australia\\
}

\maketitle

\noindent{\bf Abstract}

\textit{Reviewing radiology reports in emergency departments is an essential but laborious task. Timely follow-up of patients with abnormal cases in their radiology reports may dramatically affect the patient's outcome, especially if they have been discharged with a different initial diagnosis. Machine learning approaches have been devised to expedite the process and detect the cases that demand instant follow up. However, these approaches require a large amount of labeled data to train reliable predictive models. Preparing such a large dataset, which needs to be manually annotated by health professionals, is costly and time-consuming. 
This paper investigates a semi-supervised transfer learning framework for radiology report classification across three hospitals. The main goal is to leverage both vastly available clinical unlabeled data and already learned knowledge in order to improve a learning model where limited labeled data is available.
%The main goal is to leverage clinical unlabeled data in order to augment the learning process where limited labeled data is available. 
%This paper investigates the use of clinical unlabeled data in a semi-supervised learning pipeline using a deep learning approach for classifying radiology reports across three hospitals with different focus domains. 
%In order to cope with the lack of labeled data in some target hospitals, a transfer learning technique is applied to exploit the knowledge from source hospitals that provide abundant training data. 
%To further improve the classification performance, we also integrate a transfer learning technique into the semi-supervised learning pipeline. 
Our experimental findings show that (1) convolutional neural networks (CNNs), while being independent of  any problem-specific feature engineering, achieve significantly higher effectiveness 
%\hl{(reaching a top 0.9335 F-score)} 
compared to conventional supervised learning approaches, (2) leveraging unlabeled data in training a CNN-based classifier reduces the dependency on labeled data by more than 50\% to reach the same performance of a fully supervised CNN,
%leveraging unlabeled data in training a CNN-based classifier reduces the dependency on labeled data by more than 50\%,
%leveraging unlabeled data in training a CNN-based classifier reduces the amount of labeled data \hl{(30\%-50\% of the whole labeled set for each hospital)} required  to reach the same performance in a fully supervised setting
 and (3) transferring the knowledge gained from available labeled data in an external source hospital significantly improves the performance of a semi-supervised CNN model 
 %\hl{(reached more than 0.99 recall and 0.96 F-Score, which are at least 5\% and 3\% improvements over their fully supervised counterparts, respectively)}
 over their fully supervised counterparts
 in a target hospital.
}
\section*{Introduction}
\label{sect:introduction}

Emergency departments (EDs) in hospitals are usually overcrowded by patients with various severity of health problems~\cite{lucini2017text}. Prompt diagnosis and therapeutical decisions in such streaming environment might not be always based on the best available expert opinions. For example, patients with potential limb fractures may be treated by only examining the radiology images and prior to the availability of reports written by the radiologists. The manual reconciliation of the initial diagnosis with the formal radiology report usually occurs after the patient has been discharged from the ED~\cite{koopman2015automated}. As a result, timely follow-up of patients with abnormalities in their reports is a critical task.
% that can be affected by the large number of the patients in EDs and the time-consuming procedure of manual reviewing of the reports.

Natural Language Processing (NLP) in conjunction with Machine Learning (ML) has been widely applied to facilitate manual information extraction and classification of clinical documents~\cite{uzuner2010extracting,kholghi2015active,koopman2015automated,karimi2017automatic}. However, supervised ML approaches require a large number of labeled data to effectively capture useful information from clinical text and build robust classifiers. Manually annotating such dataset incurs considerable expenses, especially in the clinical domain as it requires significant efforts from highly qualified and busy health professionals.
%The incurred cost is a major obstacle to effective clinical data analysis. 

Semi-supervised learning (SSL)~\cite{hady2013semi} and transfer learning~\cite{pan2010survey} are feasible alternatives to standard supervised machine learning approaches to alleviate the manual annotation cost. SSL approaches incorporates the information of the unlabeled data into the learning process as a solution for dealing with scarcity of the labeled
data. Another way to minimize the workload of the manual annotation and maximize the classification performance is transferring the knowledge gained from available labeled data from one hospital (source) to a similar task at a different hospital (target). 
These approaches have been successfully applied to many real-world applications, where manual data labeling is a labour-intensive and expensive task. Examples include sentiment
analysis~\cite{ortigosa2012approaching}, pharmacogenomics and personalized medicine~\cite{xu2013semi}, cancer case management~\cite{garla2013semi}, email classification~\cite{kiritchenko2011email}, language translation~\cite{ueffing2007semi}, named entity recognition~\cite{hassanzadeh2013two}, clinical concept extraction~\cite{lv2014transfer}, and data stream classification~\cite{kholghi2011active}. 
%\hl{(These example are only for SSL. Add some for transfer learning)}

%Semi-supervised learning (SSL) and transfer learning approaches have been devised to expedite the deployment of ML approaches for such automation tasks by employing huge volume of unlabelled data and transferring the knowledge learned in a different domain.

In this paper, we study a combined semi-supervised transfer learning approach for abnormality detection from radiology reports across three major hospitals in Australia with different demographic characteristics (i.e., adult, children, and mixed general hospitals). We investigate the effect of unlabeled data in reducing the dependency on labeled data
%boosting the learning effectiveness, where training data is limited, 
using self-training as our SSL approach. Self-training follows an iterative process in which an initial model that is build over a small set of labeled data uses its own predictions on unlabeled data for further learning in successive iterations.
%Self-training is an iterative process in which an initial learning model is built using a small set of labeled data and be further augmented using unlabeled data in successive iterations. 
Furthermore, in order to improve the effectiveness of the learning model, we explore a transfer learning approach that leverages the knowledge gained from available labeled data in an external source hospital.
%An empirical comparison of the state-of-the-art conventional supervised and deep learning approaches are conducted in order to select an effective algorithm for building the initial learning model.
We specifically address the following questions in this paper:

\textbf{RQ1.} Which conventional supervised or deep learning approach can build a more effective learning model for clinical document classification?

\textbf{RQ2.} To what extent does semi-supervised learning help to deal with the scarcity of labeled data to build an effective learning model for clinical document classification?

\textbf{RQ3.} How can the knowledge from already built models from an external source hospital be used in the SSL framework to further increase the performance of the classification task?
\section*{Related Work}
\label{sect:related_work}

%Semi-supervised learning is an efficient way to cope with the lack of labeled data and the overall efforts, in terms of cost and time, for generating them.
The two primary areas that relate to this study are: (i) semi-supervised learning, and (ii) transfer learning for document classification.
Semi-supervised learning is an efficient way to cope with the lack of labeled data, where an abundance of unlabeled data is available at low cost. 
%However, manual annotation of a large number of unlabeled data is very costly and time-consuming. 
Such approaches aim to maximize the effectiveness of the learning model and minimize the manual annotation effort by incorporating the information from unlabeled data into the learning process~\cite{hady2013semi}. Many semi-supervised approaches have been developed for classification tasks, for example, self-training~\cite{nigam2000analyzing}, co-training~\cite{blum1998combining}, transductive support vector machines~\cite{joachims1999transductive}, graph-based methods~\cite{zhu2003semi}, and Expectation-Maximization (EM) with generative mixture models~\cite{nigam2000text}. Self-training is a commonly used approach, in which a classifier is first built on a small set of labeled data. Then, in an iterative process, those unlabeled samples with their predicted labels, for which the current classifier has higher confidence about their labels, are added to the labeled set. The extended labeled set is used to retrain and update the underlying classifier in each iteration. In this study, our approach conforms to the main concept of the self-training algorithm.
%\hl{[I think we can put self-training algorithm here after explanation]}
Although semi-supervised learning has been widely used in several domains (as listed above), its application in the realm of medical text analytic was limited to a small number of document classification and information extraction tasks~\cite{garla2013semi,dligach2015semi,beaulieu2016semi,wang2012extracting}.

For clinical document classification, Garla et al~\cite{garla2013semi} developed a semi-supervised framework using Laplacia SVM to recognize potentially malignant liver lesions from CT, MRI, and ultrasound reports. They showed that semi-supervised learning using Laplacian SVM significantly improved the effectiveness of the clinical text classification  (Macro-F1 0.773) when compared to a supervised SVM (Macro-F1 0.741). Although they presented encouraging results by employing unlabeled data, their classifier trained on a set of complex rule-based features narrows the application of such approach to a particular type and style of reports from a specific institution. In contrast, the proposed semi-supervised framework in this work employs simple vector representation of documents and is validated over multi-institutional data.

Another approach addressed drug-drug interaction (DDI) extraction from medical literature using distant supervision~\cite{li2016topic}. They presented a Bayesian model in a knowledge-driven distant supervision setting that incorporates available clinical knowledge resources. Drugbank and DailyMed resources were used as sources of unlabeled drug-drug relationships. Sentences that include components that have already-known relations were assumed to be the evidence of true relationships. This provides a semi-supervised learning approach that employs unlabeled data, although from a different perspective than our approach.
%This resulted in a semi-supervised learning approach that employs unlabeled data, although from a different perspective than our approach.  

Semi-supervised learning has been also used for medical information extraction tasks; Dligach et al\cite{dligach2015semi} presented a semi-supervised algorithm based on Expectation Maximization (EM) to address phenotyping tasks. They used a bag of Unified Medical Language System (UMLS) concept unique identifiers (CUIs)~\cite{bodenreider2003exploring} to represent the patients' phenotype information. They evaluated their approach over four datasets of 600 annotated patients' records. Their experiments showed that using unlabeled data improved the accuracy of the model. 
%but there was not always a positive correlation between using more unlabeled data and the performance of the model. 
Similarly,  Beaulieu-Jones et al~\cite{beaulieu2016semi} presented a semi-supervised learning approach based on denoising autoencoders (DA) for phenotype stratification that was evaluated on multiple simulation models.

Another solution to cope with the lack of training data is transfer learning, which uses available training data from {\em source} hospitals to augment the learning process in a {\em target} hospital.
%Transfer learning provides another solution to cope with the lack of training data in a particular hospital ({\em target}) using available training data from other hospitals ({\em source}).
%To address the limitation in acquiring labelled data for developing machine learning approaches across different institutions, we investigate transferability of models generated from one institution ({\em source}) to another ({\em target}). 
Given a source domain and its corresponding learning task, denoted by $D_s$ and $T_s$, respectively, transfer learning aims to improve a target learning task ($T_t$) in a target domain ($D_t$) by applying the knowledge learned from $T_s$ in $D_s$~\cite{pan2010survey}. The degrees of similarity between $D_s$ and $D_t$, and their corresponding tasks $T_s$ and $T_t$ lead to different transfer learning cases~\cite{pan2010survey,lu2015transfer}; two common forms of transfer learning in a supervised learning setting are: \textit{Inductive transfer learning} and \textit{Transductive transfer learning}. In the former method, it is considered that $T_s\neq T_t$ (regardless of the similarity of $D_s$ and $D_t$), while in the \textit{Transductive} method, also known as {\em domain adaptation}, $D_s \neq D_t$ and $T_s = T_t$.
Since the source and target tasks in our case study are similar (i.e., detecting abnormalities from radiology reports) but the domains are different (i.e., different patient demographic focus of the hospitals that is reflected in their reports), we employ the \textit{Transductive transfer learning} in our approach~\cite{hassanzadeh2018transferabiliy}. 
From a deep learning perspective, such knowledge transference is usually performed by transferring weights or features learned in different layers of a source model to a target model~\cite{yosinski2014transferable}. Degrees of similarity between the source and target domains' data and tasks affect the extent of the feature transference~\cite{yosinski2014transferable}. 
%In our experiment, source and target domains are different (e.g., an adult hospital as the source and a children hospital as the target) while source and target learning tasks are similar (i.e., fracture identification from X-Ray reports). 
%\hl{Given the difference in the domains but the similarity of the source and target tasks in our case study (as discussed above), the majority of features/layers from the source model is transferred to the target model.}
%\textbf{In our experiment, the majority of features/layers are transferred from source model due to the similarity of the source and target tasks and relatively limited number of layers in our neural network architecture.}

%\hl{I am not sure how to fit bevan?s paper here??}
%In general, literature on diagnostic classification of radiology reports have long shown the success of SVM classifiers. Koopman et al.~\cite{koopman2015automated} used Sequential Minimization Optimization (SMO) classifier which is a type of SVM in which training was performed according to the sequential minimal optimisation algorithm with a polynomial kernel. They experimented and analysed the effectiveness of their proposed method using 2,378 reports from three different hospitals. 

% !TEX root = main_ssltran.tex
\section*{Methodology}
\label{sect:method}

\subsection*{Dataset}
\label{sec:data}

We use a set of limb radiology reports that was collected within a one-year period from the EDs of 
Royal Brisbane and Women Hospital (RBWH), 
Royal Children Hospital (RCH), and Gold Coast Hospital (GCH)
(Ethics approval was granted by the Human Research Ethics Committee at Queensland Health to use the non-identifying data).
%\footnote{Ethics approval was granted by the Human Research Ethics Committee at Queensland Health to use the non-identifying data}. 
%A set of randomly selected reports were annotated by two expert annotators and refined by an adjudicator (inter-annotator agreement is Fleiss' kappa ($\kappa$) of 0.85). 
The radiology reports were classified as either ``normal" (i.e., those without any fracture, dislocation, presence of a foreign body, or incidental findings) or ``abnormal" (i.e., those with some fracture, dislocation, foreign body, or incidental findings) by two annotators and one adjudicator (all medical experts)~\cite{koopman2015automated}. The inter-annotator agreement between the two annotators was 0.85 Fleiss' kappa ($\kappa$), which exhibits a strong agreement.
Table~\ref{tab:regresnewdata} shows the distribution of normal and abnormal cases for each hospital's data. In addition, a combined set of 11,802 unlabeled data, available from two of the hospitals, was used in our semi-supervised approach.
The differences in the domain of these hospitals, explained by institutional demographic variations, reflected the variations in needs and services required to care for children, adults or both.

\begin{table}[H]
\small
  \centering
  \caption{Dataset statistics: radiology reports from three hospitals.}
  \label{tab:regresnewdata}
%\hspace*{-1.5cm}
  \begin{tabular}{ l  l  l  l }  % | l
 \toprule
Dataset              & \# Reports & Normal & Abnormal  \\[0.5ex] % & Unlabelled
\hline
RBWH                & 1,480     & 58\%  & 42\%      \\[0.5ex] % & 8,968 
RCH                    & 498     & 66\%  & 34\%    \\[0.5ex] % & 2,832 
GCH                    & 400     & 62\%  & 38\%   \\[0.5ex] %& -  
\bottomrule
\end{tabular}
\end{table}

\subsection*{Self-training}
\label{sec:ssl-frame}

\begin{figure*}[t]
% \hspace{-3cm}
\centering
%   \begin{minipage}{\linewidth}
  \includegraphics[width=.6\linewidth]{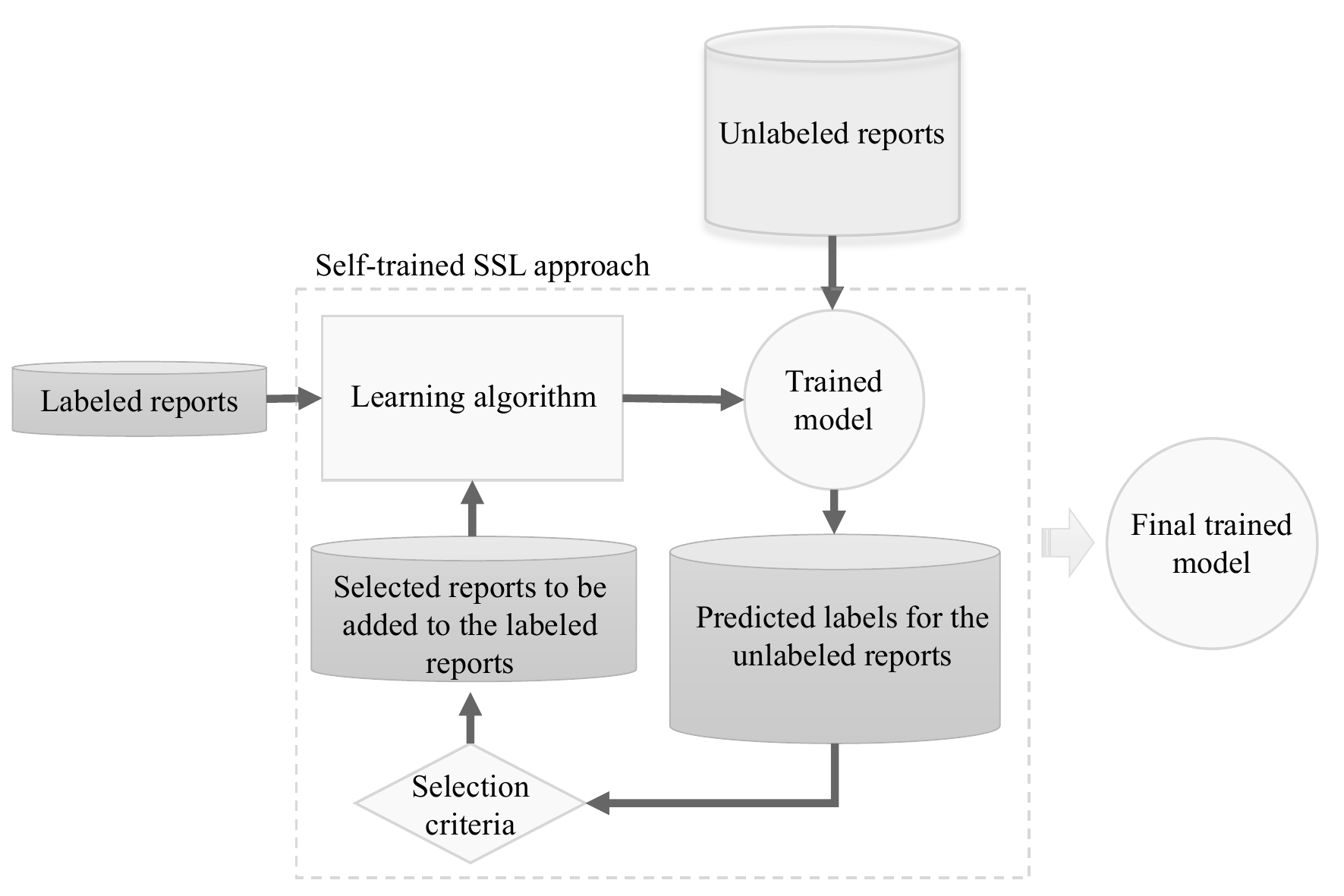} 
  %\hspace{0.5cm}
%   \qquad
%   \subfloat[][]{\includegraphics[width=.55\linewidth]{./ssl_ensemble.png} \label{fig:sslensemble}}
%   \end{minipage}
  \caption{Self-trained semi-supervised learning architecture}
  \label{fig:sslframe}
\end{figure*}

Semi-supervised learning approaches have been devised to utilize the huge amount of unlabeled data that is available in the majority of the domains demanding automation~\cite{ortigosa2012approaching,xu2013semi,garla2013semi,kiritchenko2011email,ueffing2007semi,hassanzadeh2013two,kholghi2011active}.   
%Semi-supervised methods are beneficial in machine learning tasks especially when plenty of unlabelled samples are available and easy to access but labelled data is scarce and expensive to be prepared.
One of the most common semi-supervised approaches is self-training~\cite{nigam2000analyzing}. 
%Figure~\ref{fig:sslselect} and Figure~\ref{fig:sslensemble} show the overall architecture of these two approaches. 
Figure \ref{fig:sslframe} shows the learning process in a self-training algorithm. First, an initial learning model is built using available training data (usually a small set of data). 
This model is then used to predict labels for the set of unlabeled data. The aim is to boost the performance of the model using unlabeled data through an iterative process. A selection criterion is employed to decide which subset of unlabeled samples and their predicted labels are qualified to be used for retraining the model. 
Such self-training idea is used in our combined semi-supervised transfer learning approach that is described in the following subsection.

%In order to start the self-training process from a more reliable initial model

% The ensemble-based SSL (Algorithm~\ref{alg:ensemblessl}) follows similar scenario as the self-trained approach at the beginning of the process. However, instead of training a single model, odd number of different high performing learning algorithms (three classifiers in our experiment) were used to train on the initial labelled data in order to form an ensemble of classifiers. This ensemble is then used to predict labels for the instances in the unlabelled set. 

% The final label for each instance is selected through a voting system: the final label should be given by the majority of the classifiers (having odd number of classifiers within the ensemble enables such voting scenario). The newly labelled data in conjunction with the initial labelled data was then used to train a final model using the best performing supervised learning algorithm.

%Amongst unlabelled samples, there are some that the classification model is confident about their label. These samples can be automatically labelled by the model and added to the labelled set in an iterative process, so as to provide further learning examples to the classifier. This approach is often referred to as semi-supervised learning [Ref1]. 

\subsection*{Semi-supervised Transfer Learning}
\label{sec:transfer}

%Most of ML approaches require abundant labelled data in order to train and produce a reliable learning model. In deep learning approaches such dependency to big data is even more evident in the literature\hl{CITE}. In order to cope with such shortcoming, transfer learning approaches provide a means to transfer knowledge from domains with sufficient training data to other target domain with no or limited labelled data~\cite{pan2010survey,lu2015transfer}. From a deep learning perspective, such transference is usually performed by transferring weights or features learned in different layers of a source model to a target model~\cite{yosinski2014transferable}. Degrees of similarity between the source and target domains' data and tasks support the decisions for defining the extent of the feature transference~\cite{yosinski2014transferable}. In our experiment, source and target domains are different (e.g., an adult hospital as the source and a children hospital as the target) while source and target learning tasks are similar (i.e., fracture identification from X-Ray reports). As a result, the majority of features/layers from the source model is transferred to the target model.  

In order to build a more robust initial model, we integrate a transfer learning method into the self-training process. The intuition behind such a combined framework is to leverage prior knowledge that is gained from the training data available in a source domain, in addition to the freely available unlabeled data in the target domain. The details of the self-trained transfer learning approach is described in Algorithm~\ref{alg:selfssl}. 

In the first step in Algorithm~\ref{alg:selfssl}, a model ($\Theta_i$) is trained on the available labeled data in a source domain (i.e., $\Lambda_s$). 
In the first iteration of a self-training algorithm (shown as the SSL Loop), this source model is fine-tuned using the available labeled samples in the target domain ($\Lambda_t$). Fine-tuning in this setting means tuning the already learned weights according to the new labeled data. This model is then used for predicting labels for all the samples in the unlabeled data ($\upsilon \in \Upsilon$). In the fifth step, the confidence of the model for predicting labels is assessed using a selection criterion. 
This criterion is based on the probability of the model ($P_{\Theta_i} (y|\upsilon)$): the higher the probability, the more confident the model is~\cite{sheldon2002first}. A probability threshold ($\tau$) is set to identify confidently predicted cases. 
After an empirical analysis and in order to ensure selection of samples with high precision, the $\tau$ was set to 0.99 in our approach.

{\centering
\begin{minipage}{.7\linewidth}
\begin{algorithm}[H]
\caption{Self-trained transfer learning}
\label{alg:selfssl}
\hspace*{\algorithmicindent} \textbf{Input} \\
\hspace*{\algorithmicindent}  $\Lambda_s$ \text{: set of labeled samples in the source domain}\\
\hspace*{\algorithmicindent}  $\Lambda_t$ \text{: set of labeled samples in the target domain}\\
\hspace*{\algorithmicindent}  $\Upsilon$ \text{: set of unlabeled samples in the source or target domain}\\
%\hspace*{\algorithmicindent}  $\textit{$\Theta$} \text{: classifier model}$
\hspace*{\algorithmicindent}  $\tau$ : a probability threshold used for assessing the model's confidence \\
\hspace*{\algorithmicindent} $i$ : loop's indicator ($=0$)\\
\hspace*{\algorithmicindent} \textbf{Output} \\
\hspace*{\algorithmicindent} \text{Trained model for the target task} $\Theta_t$
\begin{algorithmic}[1]
\State Train a model \textit{$\Theta_i$} on $\textit{$\Lambda_s$}$
\Procedure{SSL Loop}{}
\State Fine-tune the model \textit{$\Theta_i$} on $\textit{$\Lambda_t$}$
\State $\forall \upsilon \in \Upsilon$ predict $y$ $\gets \Theta_i(\upsilon)$
\State $\Gamma \gets$ Select a subset of $\upsilon$ from $\Upsilon$ where $P_{\Theta_i} (y|\upsilon) > \tau$ 
\State $\Gamma \gets$ $Balancing(\Gamma)$
\State $\Lambda_t \gets \Lambda_t\cup \Gamma$ % samples from step 3 to labelled set \textit{$\Lambda$} and remove them from the unlabelled set \textit{$\Upsilon$}.
\State $\Upsilon \gets \Upsilon - \Gamma$
%\State  Repeat steps 2 to 5:
\If {$Accuracy(\Theta_i) > Accuracy(\Theta_{i-1})$}  
\State $i \gets i +1$
\State \textbf{goto} 2
\Else
\State $\Theta_t \gets \Theta_{i-1}$
\State \textbf{exit} $loop$
\EndIf
%\State $j \gets \textit{patlen}$
% \emph{loop}:
%\If {$\textit{string}(i) = \textit{path}(j)$}
%\State $j \gets j-1$.
%\State $i \gets i-1$.
%\State \textbf{goto} \emph{loop}.
%\State \textbf{close};
%\EndIf
%\State $i \gets i+\max(\textit{delta}_1(\textit{string}(i)),\textit{delta}_2(j))$.
%\State \textbf{goto} \emph{top}.
\EndProcedure
\end{algorithmic}
\end{algorithm}
\end{minipage}
\par
}

In the self-training iterations, a model can be biased towards its predictions of a particular class than the rest of the classes~\cite{hassanzadeh2014load}. To avoid this problem, we use a balancing strategy. In step 6, highly distributed predicted classes are under-sampled to reach a balanced version of $\Gamma$ with the same number of samples for all classes. In a binary setting, this means keeping the same number of samples of both classes.
%a model can learn predicting a particular class more than the rest. 
In the seventh step, the newly selected and balanced set of samples (i.e., $\Gamma$) and their predicted labels are then used to augment the initial labeled data ($\Lambda \cup \Gamma$) and the model is subsequently retrained over the updated version of the training data. The performance of the retrained model (i.e., $Accuracy(\Theta_i)$) was examined (by monitoring its learning curve) to keep or roll back the model (i.e., stopping criterion to exit the loop).

\subsection*{Supervised Learning Baselines}
\label{sec:supervised}

We investigated the performance of five state-of-the-art conventional supervised learning approaches and one artificial neural network for the abnormality identification task from radiology reports. Choice of algorithms was supported by the literature and their applicability for clinical document classification tasks~\cite{karimi2017automatic,koopman2015automated, hassanzadeh2018transferabiliy}. These approaches are described in Table~\ref{tab:classifs}. 
The parameters of the models were tuned using a grid search approach.
%, where applicable.}

%~\cire{vapnik2013nature}

%\emph{Support Vector Machine (SVM)} is one of the most commonly used ML algorithms for addressing classification problems [47]. The basic idea of SVM is to represent the data instances in a high-dimensional space and then to find the optimal hyperplane that can separate categories of instances. SVM has been successfully used to classify radiology reports [Bevans].

%\emph{Na�ve Bayes (NB)} is a probabilistic classifier that is based on Bayes' theorem and conditional probabilities [53]. NB classifier applies Bayes' theorem over the features of a given instance and calculates the probability of each class. The class with the highest probability is then selected as the correct class of this instance. Na�ve Bayes classification has been previously applied to text retrieval and classification tasks including information extraction from clinical literature [49, 56, 57].

The CNN model that is devised for this study (adapted from~\cite{kim2014convolutional}) employed a word2vec model (using Skip-gram algorithm~\cite{Mikolov:2013}) that was trained on a corpus of clinical documents (approx. 5 million progress notes)~\cite{hassanzadeh2018transferabiliy}. Other hyper-parameters were tuned to optimize its performance; the final tuned values using a grid search approach were as follows: filter sizes = 10, 15; number of filters = 400; embedding model = Skip-gram on the corpus of clinical documents (mentioned above); embedding dimension = 300; learning rate = 0.0005; optimizer function = Adam; dropout rate = 0.7; batch size = 16; number of epochs = 10.

In order to provide better comparisons between the supervised approaches, all conventional classifiers in our experiment were also trained over the same vector representations that was used as the input for the CNN model. As a result, each document is represented with a vector of size 300, which equals to the embedding dimension size. This vector is an element-wise average of vectors of all constituent words in a given document.

\begin{table*}[hb]
\small
\centering
\caption{Classification algorithms and their description }
\label{tab:classifs}
\begin{tabular}{p{4cm}p{11cm}}
\hline
\textbf{Algorithm} & \textbf{Description} \\
\hline
         Support Vector Machine (SVM)      &     One of the most commonly used ML algorithms for addressing classification problems. It represents the data samples in a high-dimensional space and finds the optimal hyperplane that can separate categories of samples~\cite{vapnik2013nature}.                \\ \hline
         Na\"ive Bayes (NB)          &        A probabilistic classifier that is based on Bayes' theorem and conditional probabilities. It applies Bayes' theorem over the features of a given sample and calculates the probability of each class~\cite{maimon2005data}.             \\ \hline
          Stochastic Gradient Descent (SGD)         &        A stochastic gradient descent learning routine that supports different loss functions and penalties for classification. It updates the weights after seeing each sample based on the gradient of the loss function~\cite{bottou1991stochastic}.           \\ \hline
          Random Forest (RF)         &    An ensemble learning method that takes both bagging and random feature selection techniques to build an ensemble of tree-based classifiers~\cite{breiman2001random}.                 \\ \hline
          Logistic Regression (LR)            &        A simple classification algorithm for predicting a binary discrete variable. It uses a ``sigmoid" or ``logistic" function to predict the probability that a given example belongs to each of the classes~\cite{le1992ridge}.             \\ \hline
          Convolutional Neural Netwrok (CNN)      &      A feed-forward neural network that passes the information from the input layer through one or multiple intermediate convolutional functions to the output layer. The input is a grid-like representation of the data (e.g., a matrix of numerical representation of the constituent words of a document)~\cite{Goodfellow-et-al-2016}.          \\ \hline
\end{tabular}
\end{table*}

\subsection*{Experimental Setup and Evaluation Measures}
\label{sec:expset}

The deep learning algorithm was implemented using Keras 2.0.4~\cite{chollet2015keras} with a Theano 0.9 backend~\cite{theano}. The results were obtained using a 
%CSIRO's Bracewell 
GPU cluster. The cluster has 114 nodes, each of them has 4 NVidia Tesla P100 GPUs, 256 GB of RAM and 1TB of local SSD drive. The program was written in Python 3 and the word2vec models were generated using the Gensim library~\cite{rehurek_lrec}.
%A stratified 10-fold cross validation was used for validating and reporting the results.
%A 10-fold cross validation using all available labeled data was performed to optimize the values for the neural network' parameters (via a grid search). 
The scikit-learn implementation of conventional algorithms (Table~\ref{tab:classifs}) were used in our experiments~\cite{pedregosa2011scikit}.

%We performed 10-fold cross validation experiments to evaluate the performance of classifiers built using supervised, semi-supervised, and semi-supervised transfer learning approaches. In our evaluation, the classifier performance is measured using standard text classification measures, namely, Precision, Recall, and F1-Score:
%The performance of the conventional supervised and deep learning approaches was evaluated using 10-fold cross validation based on the standard text classification measures, namely, Precision, Recall, and F1-Score:
We performed stratified 10-fold cross validation experiments to evaluate the performance of classifiers using standard text classification measures, namely, Precision, Recall, and F1-Score:

\textit{Precision (P):} TP / (TP + FP);

\textit{Recall (R):}  TP / (TP + FN);

\textit{F1-Score (F1):} (2 * R * P) / (R + P); i.e, harmonic mean of Precision and Recall;

where true positive (TP) indicates that a model correctly identified a radiology document with a reported abnormality, false positive (FP) refers to the identification of an incorrect abnormal case, and false negative (FN) indicates that a model failed to identify an abnormality according to the ground truth data. To demonstrate statistically significant improvements
on F1-Score, we performed a paired t-test.

% !TEX root =main_ssltran.tex
\section*{Results and Discussion}
\label{sect:result}

\subsection*{Supervised Learning Performance}
\label{sec:supres}

In order to answer the \textit{RQ1}, the comparative performance of the conventional supervised ML and deep learning approaches on abnormality detection has been investigated across three hospitals (Table~\ref{tab:supres}). 
The best-performing systems in terms of Precision, Recall, and F1-Score are shown in boldface numbers. 
% and the best F1-Score across three hospitals \hl{are shaded in gray. (NO SHADING IN THE TABLE)}.

The results in Table~\ref{tab:supres} shows that Logistic Regression (LR) achieved the lowest Recall across all datasets, which led to the worst supervised system in terms of F1-Score. It should be noted that in such tasks, missing abnormal cases are more critical than misclassifying normal cases. Among the conventional supervised approaches, SVM and Naive Bayes (NB) achieved better performances in terms of F1-Score across all three hospitals.

The CNN model obtained the highest Recall across all datasets and significantly outperformed all conventional supervised approaches with the highest F1-Score of 0.9085, 0.9367, and 0.9335 across RBWH, RCH, and GCH, respectively. 
%Among the three datasets, the model built on RCH and GCH achieved higher F1-score compated to the one built on RBWH. 
Therefore, the CNN algorithm, as the best performing approach, was used in self-training and the semi-supervised transfer learning experiments to build classifiers.

% MOVE TO DISCUSSION --> Comparing the results between two datasets, the performance of CNN, SVM, and SGD on RCH dataset is comparatively higher than on RBWH. \hl{Can we tell the reason here why this is happening? For example: RCH is a smaller imbalanced dataset compared to RBWH?..}

%\cellcolor[gray]{0.8}
\begin{table}
\small
\centering
\caption{Supervised ML performances on each hospital data. Statistically significant improvements
($p$-$value<0.05$) for F1 when compared with all other supervised models are indicated by *.}%-  (full gird)
\label{tab:supres}
\begin{tabular}{llll|lll|lll}
\toprule
\multirow{2}{*}{} & \multicolumn{3}{c|}{RBWH}          & \multicolumn{3}{c|}{RCH}       & \multicolumn{3}{c}{GCH}        \\ 
                  & P      & R      & F1              & P      & R      & F1               & P      & R      & F1    \\ \hline
%SVM-bag               & 0.9109 & 0.835  & \cellcolor[gray]{0.8} 0.87   & 0.9347 & 0.9173 & \cellcolor[gray]{0.8} 0.9235  &   &   &  \\ \hline
SVM              & 0.8539	&	0.8122	&	0.8325 & 0.9366	&	0.8811	&	0.9080  &   0.9347	&   0.8810	&   0.9071  \\ \hline
%SVM              & 0.854	&	0.812	&	0.832 & \textbf{0.937}	&	0.881	&	0.907  &   0.935	&   0.881	&   0.905  \\ \hline
%SGD-bag               & 0.8969 & 0.8673 &  \cellcolor[gray]{0.8} 0.8809 & 0.9288 & 0.911  & \cellcolor[gray]{0.8} 0.918  &   &   &  \\ \hline
SGD               & 0.8575	&	0.7329	&	0.7903 & 0.9104	&	0.8276	&	0.8670  &   0.8713	&   0.7951	&   0.8315   \\ \hline
%SGD               & 0.858	&	0.733	&	0.768 & 0.910	&	0.828	&	0.839  &   0.871	&   0.795	&   0.804   \\ \hline
%NBMultiNom                & \textbf{0.7284} & 0.8754 & 0.7940           & \textbf{0.9909} & 0.636  & 0.7715     &   &   &       \\ \hline
NB               & \textbf{0.9353} & 0.7102 & 0.8074           & 0.8409 & 0.9048  & 0.8717     &  0.8049   &	0.9281 &	0.8621 \\ \hline
%NB               & \textbf{0.935} & 0.710 & 0.806           & 0.841 & 0.905  & 0.868     &  0.805   &	0.928 &	0.859 \\ \hline
%RF-bag                & 0.865  & 0.7767 & 0.8174          & 0.9532 & 0.779  & 0.8539     &   &   &       \\ \hline
RF               & 0.8508	&	0.7524	&	0.7986          & 0.9182	&	0.7552	&	0.8288      & 0.8654    &	0.8210  &	0.8426      \\ \hline
%RF               & 0.851	&	0.752	&	0.798          & 0.918	&	0.755	&	0.825      & 0.865    &	0.821  &	0.840      \\ \hline
%LR-bag                & 0.8939 & 0.8187 & 0.8532          & 0.944  & 0.7971 & 0.8622      &  &   &      \\ \hline
LR                & 0.8872	&	0.6912	&	0.7770         & 0.7003	&	0.0725	&	0.1314       &  \textbf{0.9751}  &	0.5043  &	0.6648  \\ \hline
%LR                & 0.887	&	0.691	&	0.776         & 0.700	&	0.072	&	0.128       &  \textbf{0.975}  &	0.504  &	0.659  \\ \hline
CNN               & 0.9159 & \textbf{0.9028} &  \textbf{0.9085}* & \textbf{0.9370}  & \textbf{0.9408} &  \textbf{0.9367}*   & 0.9359	&   \textbf{0.9342}    &	\textbf{0.9335}*   \\ % \hline 
\bottomrule   \\               
\end{tabular}

\end{table}

\subsection*{Semi-supervised Learning Performance}
\label{sec:semsupres}

%We now study how iteratively including samples to the labelled set () for which the classification model is very confident about their automatically assigned labels could lead to further improvement in detecting abnormalities from radiology reports. To identify these high confidence samples, two different semi-supervised scenarios, i.e., self-trained SSL, and ensemble-based SSL, are experimented. 
%According to the results reported in Table~\ref{tab:supres}, CNN achieved the highest performance on both datasets. Therefore, it is used as our learning algorithm in the self-trained SSL scenario (Figure 1(a)). 

%To build an ensemble-based Semi-Supervised Learning framework, we use two top-performing conventional supervised approaches, i.e., SVM and SGD (shown in Table~\ref{tab:supres}), along with CNN to investigate the effect of the vote-based ensemble approach in improving the classification performance.

Table~\ref{tab:sslres} presents the performance of the self-trained CNN across RBWH, RCH, and GCH. For each dataset, the self-training experiments were performed with different random portions (i.e., 30\%, 50\%, and 70\%) of the labeled set to build different initial learning models. 
The aim is to investigate the effect of using unlabeled data to learn a model that reaches at least the same performance of a fully supervised CNN model (as shown in Table~\ref{tab:supres}). 
%The aim is to investigate how using unlabeled data in the learning process can reduce the amount of required labeled data while reaching at least the same performance of a fully supervised CNN model (according to Table~\ref{tab:supres}). 
We also examined the self-trained CNN approach using 100\% of the labeled data (i.e., the whole training data) available for each dataset. This was done to investigate whether using the whole labeled set along with the unlabeled data would lead to a significant improvement in classification performance compared to a fully supervised CNN. 
The best self-trained CNN with a significant improvement over the fully supervised CNN model is indicated by an asterisk symbol and those models that reached the same effectiveness as the fully supervised CNN with no significant difference (in terms of F1-Scores) are highlighted in gray.
% that show no significant difference, i.e., $p$-$value>0.05$
%The best self-trained CNN with a significant improvement on the fully supervised CNN model is indicated by * and those models that reached at least the same F1-Score as the fully supervised CNN (without significant difference) are highlighted in gray. 
%\hl{TODO: make the following F1-score as gray highlighted in Table 4: in RBWH: 50, 70, in RCH: 30,50,70,100, in GCH: 50(?), 70 }.

By comparing the results in Table~\ref{tab:sslres} with the CNN results in Table~\ref{tab:supres}, it can be observed that a self-trained CNN model can achieve the performance of a fully supervised CNN (in terms of F1-Score) by only using a maximum of 50\% of the training data.
%The results in Table~\ref{tab:sslres} shows that using only 50\% (that is selected randomly) of the training data, a self-trained CNN achieves at least the same F1-Score (0.8994, 0.9403, and 0.9178) of the fully supervised CNN (0.9085, 0.9367, and 0.9335) across all three datasets (RBWH, RCH, and GCH), respectively. 
This demonstrates the benefit of using unlabeled data in training an effective classifier with less labeled data, which means less manual annotation effort (\textit{RQ2}). On the RCH dataset, the self-trained model reached the fully supervised performance using only 30\% of the randomly selected labeled data. When using the whole training data available for each dataset in the self-training approach, only the model built on GCH training data significantly outperformed the fully supervised CNN model (i.e., 0.9499 cf. 0.9335). The following subsection presents the results of our semi-supervised transfer learning approach in which we use labeled data available from other source hospital to further improve the classifiers' performance while addressing lack of training data in target hospitals.
%The following subsection presents the results of our semi-supervised transfer learning approach by which we tried to alleviate the dependency on labeled data even further by transferring the knowledge from a source domain to a target domain.
%This motivates the use of labeled data available from other resources (i.e., transfer the knowledge to the self-trained CNN) in order to increase the performance of the model in identifying abnormality cases from radiology reports.

%It can be observed that, on RBWH the ensemble-based approach achieved higher precision (0.926) while the self-trained SSL showed better recall (0.911), and consequently, the best F1-Score (0.9162) was for the ensemble-based SSL. On the other hand, the self-trained SSL achieved better precision (0.9473) on RCH dataset. However, the ensemble-based SSL approach performed better in terms of both Recall (0.9643) and F1-Score (0.9526). Detailed comparisons among the proposed SSL approaches and the supervised models are discussed in Section~\ref{sect:discussion}.

\begin{table}
\small
\centering
\caption{Self-trained CNN performance. Statistically significant improvements
($p$-$value<0.05$) for F1 when compared with supervised CNN model are indicated by *.}%-  (full gird)
\label{tab:sslres}
\begin{tabular}{llll|lll|lll}
\toprule
\multirow{2}{*}{} & \multicolumn{3}{c|}{RBWH}          & \multicolumn{3}{c|}{RCH}          & \multicolumn{3}{c}{GCH}           \\ 
Used Labeled Data     & P      & R      & F1              & P      & R      & F1              & P      & R      & F1  \\ \hline
30\%               & 0.8652	&   0.8835    &	0.8725   & 0.9325   &	0.9471  &	\cellcolor[gray]{0.8} 0.9379  & 0.9069    &	0.9275  &	\cellcolor[gray]{0.8} 0.9157  \\ \hline
50\%               & 0.9155   &	0.8866    &	\cellcolor[gray]{0.8} 0.8994 & 0.9403 &	0.9467  &	\cellcolor[gray]{0.8} 0.9403  & 0.9098    &	0.9275  &	\cellcolor[gray]{0.8} 0.9178  \\ \hline
70\%               & 0.9196   &	0.8804    &	\cellcolor[gray]{0.8} 0.8984 & 0.9443 &	0.9463  &	\cellcolor[gray]{0.8} 0.9429  & 0.9198    &	0.9475  &	\cellcolor[gray]{0.8} 0.9317  \\ \hline
100\%               & 0.9026    &	0.9224  &	\cellcolor[gray]{0.8} 0.9121 & 0.9295 &	0.9643  &	\cellcolor[gray]{0.8} 0.9443  & 0.9541    &	 0.9475  & \cellcolor[gray]{0.8}	0.9499*  \\
\bottomrule
\end{tabular}
\end{table}

% MOVE TO DISCUSSION --> Both SSL frameworks significantly outperformed the top-performing supervised approaches, i.e., CNN, SVM, and SGD (reported in Table X) in both datasets. The ensemble-based SSL framework improved the performance of the self-trained SSL in both datasets; but not significant improvement in the RBWH dataset. 

\subsection*{Semi-supervised Transfer Learning Performance}
\label{sect:transsl}

%In the proposed SSL framework, it was assumed that the availability of in-house unlabelled data should be satisfied in order to be able to leverage this framework. However, it is also possible to leverage unlabelled data from outside a target organisation.

In the semi-supervised transfer learning experiments, we selected RBWH as the source domain, mainly due to the provision of more labeled data, and the other two hospitals as the target domains where limited labeled data is available. 
%For the combined approach, we selected the labeled set of the RBWH dataset (as it was the biggest available train set) as \textit{source}, and other two datasets (i.e., RCH and GCH) as the \textit{target}. 
Table~\ref{tab:transsslres} shows the results of the self-trained transfer learning approach on the RCH and GCH datasets. In addition to examining different random portions (30\%, 50\%, 70\%, and 100\%) of the target labeled sets, we also examined the case that no labeled data (i.e., 0\%) is available from target datasets for the learning process. The intuition is to show the effect of knowledge transferred from other source hospitals  without any available labeled data in the target domain. The best performing settings are highlighted in boldface, which are also significantly better than the fully supervised result.

As shown in Table~\ref{tab:transsslres}, the self-trained transfer learning approach (F1-Score = 0.9744 and 0.9643) significantly outperformed
the fully supervised CNN model (F1-Score = 0.9367 and 0.9335) on both RCH and GCH datasets, while only
using 70\% of the randomly selected labeled data from each dataset (\textit{RQ3}). Comparing the results in Table~\ref{tab:transsslres} with the performance of self-trained CNN in Table~\ref{tab:sslres} shows that using labeled data from  other source hospitals along with the unlabeled data from the target hospital leads to further significant improvements (indicated by $\dagger$ in Table~\ref{tab:transsslres}).
%As shown in Table \ref{tab:transsslres}, the self-trained transfer learning approach achieved 0.9744 and 0.9643 F1-Scores on RCH and GCH, respectively. These significantly outperformed the fully supervised CNN model while only using 70\% of randomly selected labeled data from each hospital's dataset. 
%\hl{NOTE: I think the result for 100\%  and 50\% in Table 5 is also significant. Please double check. }

\medskip

\begin{table}[H]
\small
\centering
\caption{Self-trained transfer learning performance. Statistically significant improvements
($p$-$value<0.05$) for F1 when compared with supervised CNN model are indicated by * and when compared with the self-trained CNN are indicated by $\dagger$.}%-  (full gird)
\label{tab:transsslres}
\begin{tabular}{llll|lll}
\toprule
\multirow{2}{*}{}  & \multicolumn{3}{c|}{RCH}          & \multicolumn{3}{c}{GCH}           \\ 
Used Labeled Data     & P      & R      & F1              & P      & R      & F1  \\ \hline
% With Fine-tuning
% 0\%              & 0.9712	&   0.9522    &	0.9597  & 0.9480    &	0.9275  &	0.9362  \\ \hline                  
% 30\%             & 0.9721	&   0.9522    &	0.9603  & 0.9771    &   0.9075  &	0.9381  \\ \hline
% 50\%             & \textbf{0.9771}   &	\textbf{0.9581}  &	\textbf{0.9660}  & \textbf{0.9794}	&   0.9275	&   0.9519  \\ \hline
% 70\%             & 0.9709	&   0.9529	&   0.9599  & 0.9586	&   \textbf{0.9542}	&   \textbf{0.9543*}  \\ \hline
% 100\%            & \textbf{0.9771}   &	0.9522  &	0.9628  & 0.9612	&   0.9408	&   0.9498  \\
% With Instance-initialisation
0\%              & 0.9561 	&	 0.9824 	&	 0.9678  & 0.9378 	&	 0.9738 	&	 0.9548  \\ \hline                  
30\%             & 0.9542 	&	 0.9526 	&	 0.9515  & 0.9239 	&	 0.9604 	&	 0.9398  \\ \hline
50\%             & 0.9405 	&	 0.9941 	&	 0.9660  & 0.9480 	&	 0.9671 	&	 0.9553$\dagger$  \\ \hline
70\%             & \textbf{0.9570} 	&	 \textbf{0.9941} 	&	 \textbf{0.9744*$\dagger$}  & \textbf{0.9570} 	&	 \textbf{0.9738} 	&	 \textbf{0.9643*$\dagger$}  \\ \hline
100\%            & 0.9552 	&	 0.9941 	&	 0.9739$\dagger$  & 0.9312 	&	 0.9738 	&	 0.9516  \\
\bottomrule
\end{tabular}
\end{table}

% \subsection*{Error Analysis}
% \label{sect:erranls}

In order to better understand the behaviour of our learning model, we study a number of misclassified reports by the semi-supervised transfer learning model. The classification errors regarding misclassifying abnormal reports as normal were mainly related to reporting and linguistic diversity in the clinical domain that makes such document classification task even more challenging. For example, a report may contain a statement like ``no bony abnormality", however, there may also be a speculative statement that refers to a ``slight lateral deviation of the patella". This introduces a complex and challenging case for the model to automatically detect the reported abnormality. Other similar observed cases included conditional propositions (e.g., ``If clinical suspicion persists for a fracture, further evaluation with CT is advised.").

%In order to better understand the behaviour of the semi-supervised transfer learning model on abnormality detection, we explored a number of  reports misclassified by this model (Table~\ref{tab:misclass}). %shows two representative samples of misclassified reports. 
%The first row shows an example of a report contains abnormality according to manual annotation by experts, but classified as normal by the model. It can be observed that ``no bony abnormality" was detected in the ``FINDINGS" section of the report. However, there is a speculative statement that refers to a ``slight lateral deviation of the patella". This introduces a complex and challenging case for the model to automatically detect the reported abnormality. The next example, in the second row of Table~\ref{tab:misclass}, presents a normal case that is classified as abnormal by the model. Similarly, ``no underlying bony fractures" were reported by the radiologist. However, a conditional proposition suggests further examination. Such statements might lead to a misunderstanding by the model in similar cases.

These problems, known as ``ambiguity" in natural language processing, could be challenging for both human and machine to detect. Clinical free text resources, such as radiology reports, are often unstructured, ungrammatical, fragmented, and exhibit ambiguities that need to be effectively tackled when being automatically processed. As a result, employing more sophisticated NLP features is warranted in order to minimize such misclassification errors.
%Such statement might cause the model to predict an abnormality for this case, and similar reports.

A limitation of our experiment is related to the availability of unlabeled data from all hospitals. 
%A limited set of unlabeled data (i.e., 11,802 reports) was used in our semi-supervised approach that was sourced from two hospitals and was biased towards one of them (76\% provided from RBWH vs. 24\% from RCH). It can be argued that 
Despite only having unlabeled data from two of the three hospitals, our transferred learning model exhibited considerable flexibility in not only employing the source labeled model but also getting advantages of unlabeled data regardless of their domains and reporting style. Further investigation is warranted to understand the effects of quality and quantity of the unlabeled data on our approach.

\section*{Conclusion}
\label{sect:conclusion}

This paper presented a combined semi-supervised transfer learning approach to improve the performance of the abnormality detection from radiology reports. The information embedded in the unlabeled data was employed in the learning process in order to address the lack of labeled data. %Additionally, the potential use of the labeled data available from external source hospital for the classification tasks in target hospitals was studied by combining a transfer learning approach with a semi-supervised technique.
Furthermore, the potential of the labeled data
available from other source hospitals was studied in order to augment the classification performance .

The results of our empirical investigation highlighted the key role of the combined semi-supervised transfer learning approach in dealing with the scarcity of labeled data and improving the classification performance. Our results suggested that such combined approach reduces the amount of labeled data required for training an initial learning model, which can further translates into less manual annotation effort.

%The results of our empirical investigation highlighted the benefits of semi-supervised self-training approach in reducing the dependency on the labeled set to achieve at least the same performance as an effective supervised approach (i.e., CNN). Our findings suggested that combining transfer learning with the semi-supervised approach not only reduces the amount of labeled data for training, but also significantly improves the performance of the classification task compared to a fully supervised approach. 
%The semi-supervised approach studied here only use randomly selected samples to build initial learning model. 
%The future work can be directed towards exploring more effective sample selection techniques in order to form a more representative initial labeled set for semi-supervised learning.

The future work will explore more effective sample selection techniques for both generating the initial labeled set in the semi-supervised learning process and for selectively transferring informative samples from a source domain. Furthermore, we will also examine the effects of language styles and conventions used in writing the reports across institutions by using local hospital's unlabeled data instead of the combined unlabeled dataset.

\makeatletter
\renewcommand{\@biblabel}[1]{\hfill #1.}
\makeatother

\bibliography{mybibfile}

\bibliographystyle{unsrt}
%\begin{thebibliography}{1}
%\setlength\itemsep{-0.1em}
%
%\bibitem{ref1}
%Pryor TA, Gardner RM, Clayton RD, Warner HR. The HELP system. J Med Sys. 1983;7:87-101.
%\bibitem{ref2}
%Gardner RM, Golubjatnikov OK, Laub RM, Jacobson JT, Evans RS. Computer-critiqued blood ordering using the HELP system. Comput Biomed Res 1990;23:514-28.
%
%
%
%\end{thebibliography}

\end{document}